\documentclass[11pt]{article}

\usepackage[preprint]{acl}

\usepackage[T1]{fontenc}

\usepackage[utf8]{inputenc}

\usepackage{microtype}
\usepackage{times}
\usepackage{latexsym}
\usepackage{url}
\usepackage[T1]{fontenc}
\usepackage{amsmath}
\usepackage{amssymb}
\usepackage{mathtools}
\usepackage{amsthm}
\usepackage{multicol}
\usepackage{multirow}
\usepackage{longtable}
\usepackage{algorithm}
\usepackage{algorithmic}

\usepackage[most]{tcolorbox}
\usepackage[capitalize,noabbrev]{cleveref}
\usepackage{array}
\usepackage{pbox}
\usepackage{enumitem}
\usepackage{rotating} 
\usepackage{tabularx} 
\usepackage{caption} 
\usepackage{diagbox}
\usepackage{float}
\usepackage{wrapfig}
\usepackage{siunitx}
\usepackage{listings}
\usepackage{tcolorbox}
\usepackage{booktabs}
\usepackage{colortbl}
\usepackage{xcolor}
\usepackage{tikz}
\usepackage{inconsolata}

\usepackage{graphicx}

\title{SHAPE of Thinking: \\Mathematical Heuristics for Analyzing and Improving LLM Reasoning}

\author{
    \textbf{Jonghyun Song\textsuperscript{1}},
  \textbf{Sangjun Song\textsuperscript{1}},
  \textbf{Minjae Oh\textsuperscript{1}},
  \textbf{Haesung Pyun\textsuperscript{1}},
  \textbf{Sungsik Lee\textsuperscript{2}},
  \textbf{Yohan Jo\textsuperscript{1}\thanks{Corresponding author}}
\\
  \textsuperscript{1} 
  Graduate School of Data Science, Seoul National University
  \\
  \textsuperscript{2}Department of Mathematics Education, Seoul National University
\\
    \texttt{\{hyeongoon11, ssangjun706, kosair, haesung.pyun, lss1234567@snu.ac.kr, yohan.jo\}@snu.ac.kr}
}

\begin{document}
\maketitle

\begin{abstract}
Large language models (LLMs) achieve strong performance on mathematical reasoning benchmarks, yet final-answer accuracy reveals little about the strategies they use.
We introduce \texttt{SHAPE} (Structured Heuristic Analysis for Problem-solving Evolution), a framework grounded in mathematics-education research that characterizes local mathematical actions in reasoning trajectories using a problem-independent heuristic taxonomy.
Using \texttt{SHAPE}, we examine two previously hypothesized phenomena in LLM reasoning.
First, we revisit the hypothesis that LLMs reuse learned solution techniques under problem perturbations.
Across six models on MATH-Perturb, perturbations requiring a different solution approach yield lower accuracy but greater heuristic shifts than those preserving it.
Second, we characterize mode-seeking in RLVR through mathematical heuristic profiles.
Across four base--RL pairs, successful RL solutions overlap substantially with base-model profiles.
We further introduce Heuristic-Augmented Plan-GRPO (HA-Plan-GRPO), which guides RLVR planning through heuristic selection and applicability reasoning.
Across three models and five evaluation sets, it improves mean Avg@16 by up to 7.0\% absolute and mean Pass@16 by up to 2.4\% absolute over planning-only RLVR.
Our code is available at \href{https://github.com/holi-lab/shape-of-CoT}{https://github.com/holi-lab/shape-of-CoT}.
\end{abstract}

\section{Introduction}
Recent large language models (LLMs) achieve strong performance on mathematical reasoning benchmarks, often by generating long Chain-of-Thought (CoT) trajectories~\citep{hendrycks2021measuring, wei2022chain, jaech2024openai, yang2025qwen3}. 
However, final-answer accuracy offers limited insight into the mathematical strategies a model employs. 
Existing CoT analyses capture useful surface properties---length~\citep{wu2025more,su2025between}, self-revision markers such as ``wait'' or ``aha''~\citep{guo2025deepseek}, broad reasoning episodes such as planning and verification, ~\citep{li2025understanding, li2025schoenfeld}, or structural patterns such as graphs and trees~\citep{marjanovic2026deepseekr}. 
However, such work only uses coarse labels, providing limited information about the specific mathematical actions used within a solution.
Planning, for example, may involve strategic actions such as dividing into cases, introducing an auxiliary object, or working backward from the goal.
Distinguishing these actions through a common vocabulary with consistent definitions across problems would enable systematic comparisons of strategy use, providing a principled basis for diagnosing reasoning failures and improving mathematical reasoning performance.

\begin{figure*}[t]
\centering
\includegraphics[width=1.0\linewidth]{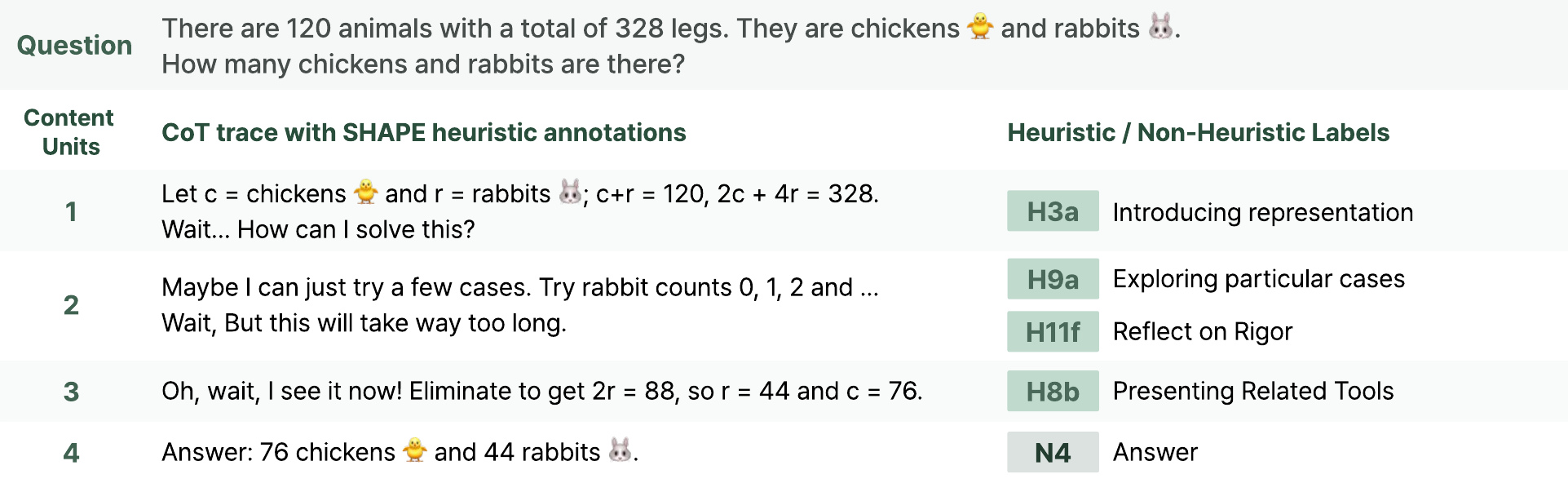}
\caption{Illustrative \texttt{SHAPE} annotations on a synthetic CoT trajectory. Each row pairs a content unit with its heuristic or non-heuristic labels. Unit~2 illustrates multi-label annotation with H9a (exploring particular cases) and H11f (reflection on rigor), while unit~4 receives the non-heuristic final-answer label N4.}
\label{fig:tagging_example}
\end{figure*}

Mathematics education research has long studied purposeful problem-solving actions, termed \emph{heuristics}~\citep{favier2024heuristics, koichu2007heuristic, rott2014rethinking}.
Heuristics can be used to analyze LLM-generated CoT trajectories and characterize the mathematical actions a model expresses during problem solving.
To that end, we introduce \texttt{SHAPE} (Structured Heuristic Analysis for Problem-solving Evolution), a framework that labels LLM reasoning as a sequence of heuristics.
To enable consistent comparison across problems and models, we organize established heuristics into eleven categories applicable across mathematical domains.
We annotate each \emph{content unit}---a span representing a coherent problem-solving step---with one or more heuristic labels.
To cover all content units, we label simple computations or other non-strategic steps as non-heuristic steps such as technical performance or literal repetition.
Using these annotations, \texttt{SHAPE} measures how heuristic activity is distributed within trajectories and how its composition varies across models and post-training conditions.
We implement \texttt{SHAPE} through an automated pipeline that scales heuristic tagging to various trajectories across models.
Figure~\ref{fig:tagging_example} illustrates these annotations on a synthetic CoT trajectory.

We use \texttt{SHAPE} to examine two previously hypothesized phenomena in LLM reasoning, on how reasoning differs with different problems and RL post-training.
First, we investigate whether models change their mathematical actions when superficially similar problems require different solution approaches.
MATH-Perturb~\citep{huang2025math} contains simple and hard perturbations that both make minimal changes to problem wording, but only hard perturbations require a different solution approach.
Prior work identifies inappropriate reuse of learned solution techniques as a failure mode under hard perturbations.
Across six models, however, hard perturbations yield lower accuracy but greater shifts in heuristic composition than simple perturbations.
Second, we examine the hypothesized mode-seeking behavior of reinforcement learning with verifiable rewards (RLVR)~\citep{guo2025deepseek, yue2025does}, in which RL concentrates on a subset of strategies the base model already uses.
\texttt{SHAPE} brings this hypothesis to the level of mathematical heuristic usage.
Across four base--RL pairs, successful RL solutions substantially overlap with base-model heuristic profiles, while coverage of the base-model distribution varies across pairs.
Olmo-3-7B-Think-RL-Zero shows the clearest pattern consistent with mode-seeking, combining high overlap with the lowest coverage, whereas the other pairs cover a broader range of base-model profiles.
These findings characterize how successful RL solutions reuse and concentrate heuristic patterns already represented in the base model.
We further apply \texttt{SHAPE} to training by integrating its heuristic taxonomy into planning during RLVR.
We propose Heuristic-Augmented Plan-GRPO (HA-Plan-GRPO), which prompts models to select a relevant mathematical heuristic and explain its applicability before carrying out each major reasoning step during training.
HA-Plan-GRPO improves mean Avg@16 by up to 7.0\% and mean Pass@16 by up to 2.4\% over planning-only RLVR across three models.

Together, our findings show that mathematical heuristics provide a common vocabulary for analyzing and improving LLM reasoning. 
Our analyses reveal differences in mathematical actions across superficially similar problems and characterize how successful usage patterns change with post-training. 
Providing these heuristics into planning improves performance under RLVR, improving their generation during training.

\section{Related Work}

\paragraph{Chain-of-thought analysis.}
A growing body of work analyzes Chain-of-Thought (CoT) trajectories beyond final-answer accuracy through reasoning length, self-revision markers, cognitive episode labels, and graph-based structure~\citep{wu2025more, guo2025deepseek, li2025schoenfeld, marjanovic2026deepseekr, jiang2025makes, xiong2025mapping, zhang2025llmsreallyneed10}.
For example, ThinkARM identifies functional episodes such as analysis, exploration, and verification~\citep{li2025schoenfeld}.
\texttt{SHAPE} focuses on the mathematical actions expressed within such episodes, distinguishing moves such as case decomposition, working backward, and introducing auxiliary objects.
Studies of CoT faithfulness show that verbalized reasoning need not faithfully reflect internal computation~\citep{arcuschin2025chain, lanham2023measuring}; our annotations therefore characterize mathematical actions expressed in observable traces.

\paragraph{Mathematical heuristics and strategy-based evaluation.}
Mathematical problem-solving research has developed heuristic frameworks for describing actions such as simplifying a problem, exploring particular cases, changing representations, and checking an argument~\citep{polya1945solve, schoenfeld1985mathematical, koichu2007heuristic, rott2014rethinking, favier2024heuristics}.
Recent LLM studies apply P\'olya-inspired methods to tutoring and evaluation~\citep{lee2026llamapolya}, and assess strategy diversity through problem-specific solution families~\citep{yang2026beyond}.
\texttt{SHAPE} organizes mathematical heuristics into a fixed, problem-independent taxonomy.
Whereas problem-specific solution families describe different ways to solve a given problem, this shared vocabulary supports comparisons of local mathematical actions across problems and models.

\paragraph{Post-training and heuristic guidance for reasoning.}
RLVR improves mathematical reasoning through outcome-based rewards~\citep{guo2025deepseek, lambert2024tulu, yang2025qwen3, wang2025nemotron}.
Studies report improved sampling efficiency and reduced reasoning diversity~\citep{yue2025does, zhao2025echo, wu2025invisible, dang2025assessing, matsutani2025rl}, while prolonged training can expand reasoning boundaries~\citep{liu2025prorl}.
Related methods use planning-based training~\citep{jiao2024learning} or rewards for aligning predicted and reference skill sequences~\citep{he2026toward}.
\texttt{SHAPE} analyzes post-training through heuristic-frequency profiles and uses the same taxonomy to guide planning during RLVR.

\section{\texttt{SHAPE}: Framework and Pipeline}
\label{sec:method}

In this section, we present \texttt{SHAPE} (Structured Heuristic Analysis for Problem-solving Evolution), a framework for analyzing the mathematical actions expressed in LLM-generated CoT trajectories. 
We first describe its foundations in mathematics-education research and the heuristic taxonomy (\S\ref{sec:theoretical_framework}), then define heuristic sequences and frequency distributions for representing heuristic usage (\S\ref{sec:metrics}), and finally introduce the automated annotation pipeline (\S\ref{sec:pipeline}).

\subsection{Theoretical Background}

\begin{table}[t]
\centering
\small
\setlength{\tabcolsep}{3pt}
\renewcommand{\arraystretch}{1.05}
\begin{tabularx}{\columnwidth}
{@{}l>{\raggedright\arraybackslash}X@{}}
\toprule
Code & Category \\
\midrule
H1 & Changing the register of semiotic representation \\
H2 & Cognitive reinterpretation \\
H3 & Introduce symbolic representation, formalization, and structural augmentation \\
H4 & Problem classification / rephrase the problem and goal \\
H5 & Wishful thinking (simplify / reduce the problem and conditions) \\
H6 & Explicit case analysis, decompose into subproblems \\
H7 & Arguing by contradiction \\
H8 & Analogy and presenting related theorems \\
H9 & Experimental and pattern exploration \\
H10 & Thinking from the end to the beginning (working backward) \\
H11 & Verification and looking back \\
\midrule
N & Non-heuristics (repetition, routine computation, unrelated statements, and final-answer statements) \\
\bottomrule
\end{tabularx}
\caption{Heuristic and non-heuristic categories used in \texttt{SHAPE}.}
\label{tab:heuristic-families}
\end{table}

\begin{figure*}[t]
    \centering
    \includegraphics[width=\linewidth]{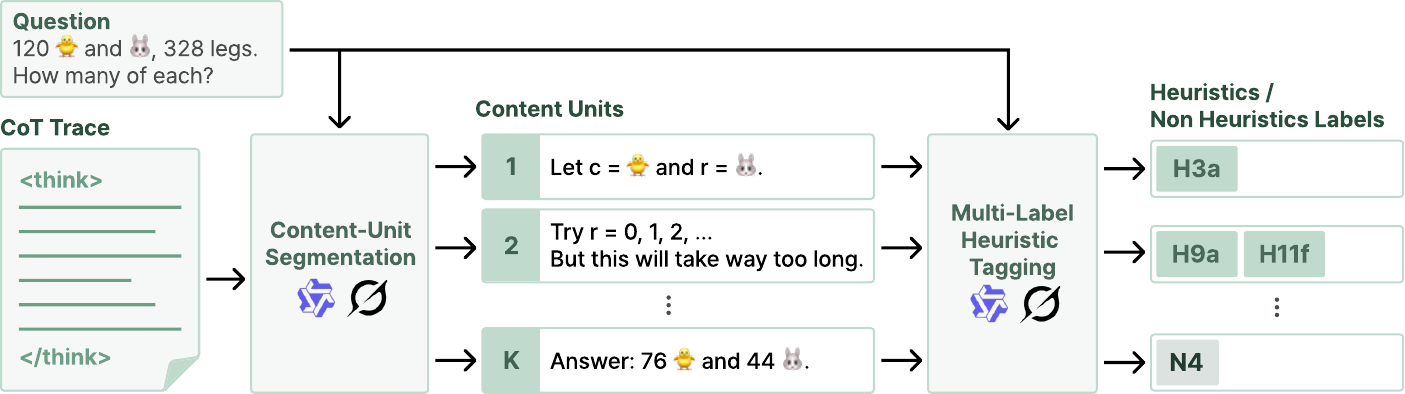}
    \caption{
Overview of the automated \texttt{SHAPE} annotation pipeline. Given a question and a chain-of-thought (CoT) trace, the pipeline segments the trace into content units \(1,\ldots,k\) and assigns one or more heuristic or non-heuristic labels to each unit. Both stages use Grok or Qwen as annotators.
    }
    \label{fig:pipeline}
\end{figure*}
\label{sec:theoretical_framework}
\texttt{SHAPE} builds on mathematics-education research that studies reasoning through observable records produced during problem solving.
In human studies, these records include verbalizations elicited through think-aloud protocols~\citep{ericsson2017protocol, koichu2007heuristic, carlson2005cyclic} and audiovisual records of problem-solving actions~\citep{favier2024heuristics}.
For LLM mathematical reasoning, a model-generated CoT trajectory can be thought of as an observable textual record of stated intermediate moves, including reformulations, transformations, verifications, and conclusions.
\texttt{SHAPE} characterizes the mathematical actions expressed in this record through a sequence of heuristic annotations.
We use \emph{heuristics} to describe purposeful moves that guide mathematical problem solving~\citep{rott2014rethinking}, including simplifying a problem, introducing auxiliary objects, dividing into cases, and working backward~\citep{polya1945solve, schoenfeld1985mathematical, koichu2007heuristic, favier2024heuristics}.
Drawing on prior taxonomies developed for different mathematical problem-solving contexts~\citep{polya1945solve, koichu2007heuristic, favier2022etude, posamentier2008problem}, we organize these heuristics into a fixed, problem-independent taxonomy of eleven families for annotating LLM-generated CoT trajectories.
Consistent label definitions provide common criteria for comparing mathematical actions across problems and models.
Table~\ref{tab:heuristic-families} lists the eleven heuristic families.
The full taxonomy, finer sublabels, and mappings to the source literature are provided in Appendix~\ref{app:taxonomy}.

\subsection{Representing Heuristic Usage}
\label{sec:metrics}
Each trajectory is segmented into \emph{content units}, the largest unbroken spans that support a particular heuristic interpretation~\citep{koichu2007heuristic}.
We assign heuristic labels based on the mathematical actions each content unit expresses in context.
For example, calculating a small case may serve to explore a pattern, check a conjecture, or carry out an already selected procedure.
A unit receives multiple labels when it uses more than one heuristic, consistent with prior analyses of concurrent heuristic use~\citep{koichu2007heuristic, favier2024heuristics}.
When no heuristic is expressed, non-heuristic labels distinguish routine computation, repetition, unrelated statements, and final-answer statements.

We represent a trajectory of $T$ content units as a heuristic sequence $\mathbf{H}=(H_1,\ldots,H_T)$, where $H_t\subseteq\mathcal{H}$ is the set of heuristic families assigned to unit $t$ and $\mathcal{H}=\{\mathrm{H1},\ldots,\mathrm{H11}\}$.
To identify the heuristics used in a solution, we represent each content unit by the set of heuristic families.
For trajectories with at least one heuristic assignment, we define the \emph{heuristic-frequency distribution} $u$ as
\begin{equation}
u(h)=
\frac{\sum_{t=1}^{T}\mathbf{1}[h\in H_t]}
{\sum_{t=1}^{T}|H_t|},
\qquad h\in\mathcal{H}.
\label{eq:heuristic-frequency}
\end{equation}
Here, $u(h)$ denotes the share of all heuristic-family assignments belonging to $h$, providing a common representation for comparing heuristic usage across problems and models.

\subsection{Automated Heuristic Annotation}
\label{sec:pipeline}
\begin{table}[!htbp]
\centering
\small
\begin{tabular}{lcc}
\toprule
Annotator & F1 (\%) & Gwet's AC1 \\
\midrule
Grok-4.1-Fast & 77.16 & 0.909 \\
GPT-5 & 71.56 & 0.859 \\
GPT-5-mini & 66.13 & 0.850 \\
Gemini-3-Flash & 65.11 & 0.860 \\
Gemini-3.1-Flash-Lite & 71.22 & 0.890 \\
\midrule
Qwen3.5-27B & 70.44 & 0.894 \\
\bottomrule
\end{tabular}
\caption{Annotator model agreement with the human gold standard.}
\label{tab:annotator_evaluation}
\end{table}

\begin{table*}[t]
\centering
\small
\setlength{\tabcolsep}{4pt}
\renewcommand{\arraystretch}{1.12}
\small
\setlength{\tabcolsep}{5pt}
\renewcommand{\arraystretch}{1.15}

\begin{tabular}{@{}l*{7}{c}@{}}
\toprule
Model
& \multicolumn{3}{c}{Pass@1 (\%)}
& \multicolumn{2}{c}{$D_{\mathrm{JS}}$ (Full)}
& \multicolumn{2}{c}{$D_{\mathrm{JS}}$ ($\leq$5)} \\
\cmidrule(lr){2-4}
\cmidrule(lr){5-6}
\cmidrule(l){7-8}
& Original & Simple & Hard
& O--S & O--H & O--S & O--H \\
\midrule
Qwen3-8B
& 96.0 & 94.0 & 84.0
& .043 & $.066^{*}$ & .160 & $.190^{*}$ \\
Qwen3-32B
& 92.0 & 90.0 & 76.0
& .047 & $.070^{*}$ & .150 & $.200^{*}$ \\
Nemotron-Cascade-8B
& 92.0 & 95.0 & 71.0
& .049 & $.072^{*}$ & .170 & $.210^{*}$ \\
Olmo-3-7B-Think-RLVR
& 97.0 & 94.0 & 76.0
& .041 & $.065^{*}$ & .165 & $.205^{*}$ \\
Qwen2.5-7B (SimpleRL-Zoo)
& 61.7 & 58.3 & 31.3
& .152 & $.253^{*}$ & .158 & $.254^{*}$ \\
Qwen2.5-14B (SimpleRL-Zoo)
& 67.0 & 69.6 & 40.9
& .197 & $.231^{*}$ & .200 & $.231^{*}$ \\
\bottomrule
\end{tabular}
\caption{
Pass@1 and heuristic-frequency divergence on MATH-Perturb.
O, S, and H denote original, simple, and hard problems.
$D_{\mathrm{JS}}$ is Jensen--Shannon divergence between frequency distributions over eleven heuristic families, using base-2 logarithms.
Full uses the entire trajectory; $\leq$5 uses the first five content units, or the entire trajectory when fewer units are available.
Means are computed over matched problems with valid O/S/H distributions, separately for each condition.
$^{*}$ denotes O--H $>$ O--S in a one-sided paired Wilcoxon signed-rank test (uncorrected $p<.05$).
For most SimpleRL-Zoo trajectories, $\leq$5 includes the entire CoT.
}
\label{tab:hard_perturb}
\end{table*}

To apply \texttt{SHAPE} at scale, we build an automated pipeline that converts raw CoT trajectories into sequences of heuristic annotations.
The pipeline consists of two stages: content-unit segmentation and multi-label heuristic tagging (Figure~\ref{fig:pipeline}).
Both stages use prompting-based annotation, with prompt templates provided in Appendix~\ref{app:annotator_prompts}.

\paragraph{Gold Standard and Annotator Model Selection}
To evaluate automated heuristic tagging, we construct a consensus-annotated gold standard of 48 CoT trajectories on MATH-Perturb~\citep{huang2025math}, comprising 7,565 sentences grouped into 1,598 content units.
The set covers both original problems and their hard perturbations, with trajectories from Qwen3-30B-A3B-Instruct, Qwen3-30B-A3B-Thinking, Qwen3-8B~\citep{yang2025qwen3}, and Nemotron-Cascade-8B~\citep{wang2025nemotron}.
Following the protocol-coding approach in mathematics-education research~\citep{koichu2007heuristic}, four authors, including a graduate researcher in mathematics education, annotated the content units and finalized the labels through consensus discussion.
Appendix~\ref{app:gold_protocol} provides the trajectory selection criteria, annotation procedures, and an illustrative example.

We evaluate six candidate annotator models on the content units using F1 averaged across content units and Gwet's AC1~\citep{gwet2008computing} over the eleven heuristic families (Table~\ref{tab:annotator_evaluation}).
We report AC1 alongside F1 because many heuristic families occur infrequently, resulting in strongly imbalanced presence--absence decisions.
Grok-4.1-Fast~\citep{grok-4.1} achieves the highest F1 and macro AC1 among the evaluated candidates.
Qwen3.5-27B~\citep{qwen3.5} provides a competitive open-weight alternative, with F1 of 70.44\% and macro AC1 close to Grok's (0.894 vs. 0.909).
We therefore use Qwen3.5-27B as the default annotator for both segmentation and tagging to support reproducible, large-scale annotation, and explicitly identify analyses annotated with Grok-4.1-Fast\footnote{An annotation run of 445 trajectories with Grok-4.1-Fast cost approximately \$100.}.
An annotator sensitivity analysis is provided in Appendix~\ref{app:annotator_sensitivity}.

\begin{table*}[t]
\centering
\small
\resizebox{\linewidth}{!}{%
\begin{tabular}{llcccc}
\toprule
Reference & Target & $N_{\text{ref}}$ & $N_{\text{target}}$
& Density & Coverage \\
\midrule
\multicolumn{6}{l}{\textit{Post-trained}} \\
Qwen3-1.7B-Base & Qwen3-1.7B-GRPO & 834 & 886 & 1.219 & 0.871 \\
Olmo-3-7B & Olmo-3-7B-Think-RL-Zero & 1229 & 1307 & 1.250 & 0.707 \\
Qwen2.5-7B & Qwen2.5-7B-SimpleRL-Zoo & 1280 & 1294 & 1.511 & 0.888 \\
DeepSeek-Math-7B-Base & DeepSeek-Math-7B-SimpleRL-Zoo & 463 & 500 & 1.525 & 0.950 \\
\midrule
\multicolumn{6}{l}{\textit{Cross-model baselines}} \\
Qwen2.5-7B & DeepSeek-Math-7B-Base & 796 & 493 & 0.594 & 0.348 \\
Olmo-3-7B & Qwen3-1.7B-Base & 71 & 66 & 0.525 & 0.437 \\
\bottomrule
\end{tabular}%
}

\caption{Density and Coverage for base--post-trained pairs and cross-model baselines in the original annotation-frequency space ($k=3$, cosine distance, successful trajectories only, aggregated across all perturbation conditions).
Density measures target overlap with base neighborhoods; Coverage is the fraction of base neighborhoods containing at least one target trajectory.
The left model defines the reference neighborhoods; $N$ counts successful trajectories.}
\label{tab:dc}
\end{table*}

\section{Analyzing Heuristic Usage}
\label{sec:analysis}
We use \texttt{SHAPE} to examine how heuristic usage varies with problem requirements and RL post-training.
Both analyses use the heuristic-frequency representation introduced in \S ~\ref{sec:metrics}.

\subsection{Heuristic Usage Shifts Despite Accuracy Drops}

\label{sec:mode-seeking-hard}
Prior work has interpreted failures under mathematical problem perturbations as evidence of pattern matching or inappropriate reuse of learned problem-solving techniques~\citep{mirzadeh2024gsm,huang2025math}.
However, final-answer accuracy alone does not fully characterize the intermediate reasoning process~\citep{xia2025evaluating}.
An accuracy drop therefore does not by itself establish that the heuristic usage expressed in a solution remains unchanged.
We use \texttt{SHAPE} to examine how the heuristic usage represented by heuristic-frequency distributions of these actions differs across problem variants.

\paragraph{Setup}
We use the 115-problem test split of MATH-Perturb, which pairs each original level-5 MATH problem with a simple and a hard perturbation.
We analyze CoT trajectories from six models: Qwen3-8B, Qwen3-32B~\citep{yang2025qwen3}, Nemotron-Cascade-8B~\citep{wang2025nemotron}, Olmo-3-7B-Think-RLVR~\citep{olmo2025olmo}, and the Qwen2.5-7B and Qwen2.5-14B SimpleRL-Zoo checkpoints ~\citep{zeng2025simplerl}.
We use Grok-4.1-Fast for content-unit segmentation and heuristic tagging.
For each matched triplet, we compare original--simple (O--S) and original--hard (O--H) trajectories using Jensen--Shannon divergence $D_{\mathrm{JS}}$ between their heuristic-frequency distributions $u$.
We compute these distributions from full trajectories (Full) and their first five content units ($\leq 5$).

\paragraph{Results}
Across all six models, hard problems have lower Pass@1 than simple problems, yet greater full-trajectory O--H divergence than O--S divergence (Table~\ref{tab:hard_perturb}).
The divergence contrast is statistically significant for all six models using both full trajectories and the first five content units (one-sided paired Wilcoxon signed-rank tests, uncorrected $p<.05$).
For Qwen3-8B, Pass@1 is $94.0\%$ on simple problems and $84.0\%$ on hard problems, while mean full-trajectory divergence is $.043$ for O--S and $.066$ for O--H.
This difference is already visible in the first five content units; Qwen3-8B yields $.190$ for O--H versus $.160$ for O--S.
Together, the full and first five content unit results show that lower accuracy on hard problems is accompanied by differences in the observed heuristic usage, including within the opening steps of the generated solutions.
Lower accuracy alone therefore does not establish unchanged heuristic usage, although inappropriate reuse of particular heuristics may still occur.

\subsection{Post-Training Concentrates Heuristic Usage}
\label{sec:mode-seeking}

Prior work suggests that RLVR exhibits mode-seeking behavior, concentrating probability mass on a narrower subset of reasoning trajectories already available to the base model, instead of discovering completely novel trajectories~\citep{yue2025does,wu2025invisible,dang2025assessing}.
However, outcome-level diagnostics such as pass@$k$ do not reveal whether this concentration extends to the composition of mathematical heuristics. 
We use \texttt{SHAPE} to examine this mode-seeking hypothesis directly at the level of heuristic composition, testing whether successful post-trained solutions concentrate in a narrower region of heuristic space already used by the base model.

\begin{figure}[t]
\centering
\includegraphics[width=\linewidth]{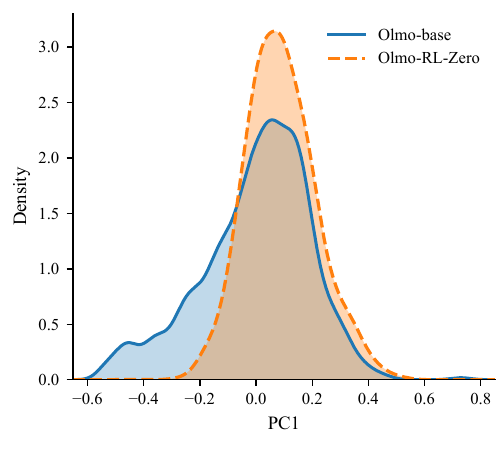}
\caption{Pooled successful heuristic-frequency profiles of Olmo-3-7B and Olmo-3-7B-Think-RL-Zero projected onto the first principal component (PC1). The RL distribution has a sharper peak near the base-model mode and a reduced left tail.}
\label{fig:pca-olmo}
\end{figure}

\begin{table*}[t]
\raggedright
\small
\setlength{\tabcolsep}{0.8pt}

\newcommand{\shapeMethod}{HA-Plan-GRPO}
\renewcommand{\arraystretch}{1.08}

\newcommand{\shapeScore}[2]{%
  \ensuremath{#1{\scriptstyle\pm#2}}}
\newcommand{\shapeBest}[2]{%
  \ensuremath{\mathbf{#1}{\scriptstyle\pm#2}}}

\begin{tabular*}{\textwidth}{@{\extracolsep{\fill}}l*{12}{l}@{}}
\toprule
\multicolumn{1}{@{}c}{Method}
& \multicolumn{2}{c}{Original}
& \multicolumn{2}{c}{Simple}
& \multicolumn{2}{c}{Hard}
& \multicolumn{2}{c}{MATH500}
& \multicolumn{2}{c}{Minerva}
& \multicolumn{2}{c@{}}{Mean} \\
\cmidrule(lr){2-3}
\cmidrule(lr){4-5}
\cmidrule(lr){6-7}
\cmidrule(lr){8-9}
\cmidrule(lr){10-11}
\cmidrule(l){12-13}
& \multicolumn{1}{c}{Avg.}
& \multicolumn{1}{c}{Pass}
& \multicolumn{1}{c}{Avg.}
& \multicolumn{1}{c}{Pass}
& \multicolumn{1}{c}{Avg.}
& \multicolumn{1}{c}{Pass}
& \multicolumn{1}{c}{Avg.}
& \multicolumn{1}{c}{Pass}
& \multicolumn{1}{c}{Avg.}
& \multicolumn{1}{c}{Pass}
& \multicolumn{1}{c}{Avg.}
& \multicolumn{1}{c@{}}{Pass} \\
\midrule

\multicolumn{13}{@{}l}{\textit{DeepSeek-Distill-Llama-8B}} \\
\addlinespace[2pt]
Base
& 58.5 & 80.9
& 56.7 & 81.7
& 41.3 & 67.0
& 62.3 & 81.8
& \textbf{15.6} & 34.2
& 46.9 & 69.1 \\

Plan-GRPO
& \shapeScore{54.4}{6.2} & \shapeScore{81.7}{1.7}
& \shapeScore{52.1}{7.0} & \shapeScore{83.2}{2.7}
& \shapeScore{35.5}{4.7} & \shapeScore{68.1}{4.3}
& \shapeScore{58.8}{5.4} & \shapeScore{82.9}{1.8}
& \shapeScore{13.0}{1.6} & \shapeScore{32.6}{2.9}
& 42.8 & 69.7 \\

\rowcolor{black!5}
\shapeMethod
& \shapeBest{63.6}{1.8} & \shapeBest{84.6}{2.7}
& \shapeBest{61.6}{0.9} & \shapeBest{84.9}{2.2}
& \shapeBest{42.4}{1.9} & \shapeBest{68.4}{1.3}
& \shapeBest{66.1}{1.6} & \shapeBest{86.6}{2.4}
& \shapeScore{15.2}{1.6} & \shapeBest{35.8}{2.8}
& \textbf{49.8} & \textbf{72.1} \\
\midrule

\multicolumn{13}{@{}l}{\textit{Qwen3-1.7B-Base}} \\
\addlinespace[2pt]
Base
& 31.6 & 67.8
& 33.5 & 70.4
& 17.7 & 51.3
& 56.6 & \textbf{85.6}
& 10.8 & 34.6
& 30.0 & 61.9 \\

Plan-GRPO
& \shapeScore{37.8}{3.3} & \shapeScore{72.5}{2.5}
& \shapeScore{37.3}{2.2} & \shapeScore{71.0}{2.2}
& \shapeScore{19.8}{1.5} & \shapeScore{50.4}{1.7}
& \shapeScore{58.9}{2.5} & \shapeScore{84.8}{1.2}
& \shapeScore{14.2}{1.5} & \shapeScore{34.8}{0.8}
& 33.6 & 62.7 \\

\rowcolor{black!5}
\shapeMethod
& \shapeBest{38.9}{1.6} & \shapeBest{73.3}{3.1}
& \shapeBest{38.4}{1.3} & \shapeBest{72.2}{0.0}
& \shapeBest{20.1}{1.3} & \shapeBest{51.6}{1.3}
& \shapeBest{60.2}{1.1} & \shapeScore{83.4}{1.0}
& \shapeBest{15.7}{1.5} & \shapeBest{35.8}{1.3}
& \textbf{34.7} & \textbf{63.3} \\
\midrule

\multicolumn{13}{@{}l}{\textit{Qwen3.5-4B-Base}} \\
\addlinespace[2pt]
Base
& 77.0 & \textbf{95.7}
& 75.5 & \textbf{95.7}
& 57.5 & 78.3
& 82.3 & 94.4
& 24.1 & 43.4
& 63.3 & 81.5 \\

Plan-GRPO
& \shapeScore{77.5}{9.0} & \shapeScore{94.3}{0.6}
& \shapeScore{76.9}{12.3} & \shapeScore{94.8}{1.2}
& \shapeScore{55.8}{10.8} & \shapeScore{77.8}{0.6}
& \shapeScore{85.2}{6.0} & \shapeScore{95.9}{0.1}
& \shapeBest{28.5}{1.4} & \shapeScore{45.6}{2.1}
& 64.8 & 81.7 \\

\rowcolor{black!5}
\shapeMethod
& \shapeBest{80.1}{3.2} & \shapeScore{94.8}{0.9}
& \shapeBest{78.0}{3.9} & \shapeScore{95.1}{0.5}
& \shapeBest{59.3}{1.7} & \shapeBest{78.6}{1.8}
& \shapeBest{86.3}{2.0} & \shapeBest{96.0}{0.2}
& \shapeScore{27.5}{2.0} & \shapeBest{46.1}{2.2}
& \textbf{66.2} & \textbf{82.1} \\
\bottomrule
\end{tabular*}
\caption{
Avg@16 and Pass@16 (\%) across five evaluation sets.
Trained variants report mean $\pm$ standard deviation across
three training seeds; Base results are point estimates.
Mean columns average the five evaluation sets with equal weights.
Light gray rows denote HA-Plan-GRPO (ours).
Bold indicates the highest mean within each model, evaluation
set, and metric.
}
\label{tab:ha_grpo_mathp}
\end{table*}

\paragraph{Setup}
We apply Density and Coverage~\citep{naeem2020reliable} to the H1--H11 heuristic-frequency vectors $u$, using the base as reference and the post-trained model as target ($k=3$, cosine distance).
For each base solution, we define a neighborhood whose radius is its cosine distance to the \(k\)-th nearest other base solution \((k=3)\). 
Density counts how many of these base neighborhoods contain each RL solution, averages this count over RL solutions, and divides by \(k\)
Coverage measures the fraction of base neighborhoods containing at least one RL solution.
Thus, Density describes overlap from the perspective of RL solutions, whereas Coverage describes how broadly they cover the observed base profiles. High Density together with low Coverage is consistent with concentration within part of the base distribution.
We compare four base--RL pairs: DeepSeek-Math-7B-Base and Qwen2.5-7B with their SimpleRL-Zoo checkpoints~\citep{zeng2025simplerl}; Olmo-3-7B with Olmo-3-7B-Think-RL-Zero~\citep{olmo2025olmo}; and Qwen3-1.7B-Base with our GRPO checkpoint~\citep{shao2024deepseekmath}.
We analyze successful trajectories on the original, simple, and hard versions of MATH-Perturb.

\paragraph{Results}
Successful RL solutions show substantial overlap with profiles observed in their base models, while the breadth of this overlap varies across model pairs.
Figure~\ref{fig:pca-olmo} shows a sharper Think-RL-Zero peak near the base peak and a reduced left tail along PC1.
In Table~\ref{tab:dc},  Olmo pair has the lowest Coverage ($70.7\%$), with Density $1.250$, a pattern consistent with overlap concentrated in part of the base model's heuristic frequency space.
The other models also exhibit high Density while covering a broader range of base-model profiles (87.1--95.0\%), indicating substantial overlap across different degrees of coverage.
In contrast, the cross-model comparisons show low Density with low Coverage, indicating weak overlap between the usage patterns of unrelated base models.
Together, these findings are consistent with a mode-seeking behavior of RLVR, in which successful solutions preferentially reuse patterns already represented in the base model~\citep{wu2025invisible, dang2025assessing, matsutani2025rl}, with the clearest concentration observed in Olmo-3.
Sensitivity to neighborhood size \(k\) is reported in Appendix~\ref{app:dc_robustness}.

\section{Heuristic-Augmented Reinforcement Learning}
\label{sec:ha-grpo-training}

In human mathematical problem solving, explicit instruction in heuristic strategies improved performance compared with practice on the same problems and worked solutions~\citep{schoenfeld1979explicit}.
Motivated by this finding, we use the heuristic vocabulary of \texttt{SHAPE} to guide solution planning during RLVR.
We evaluate its effects on task performance and conduct qualitiative analysis.

\begin{figure*}[t]
\centering
\includegraphics[width=\linewidth]{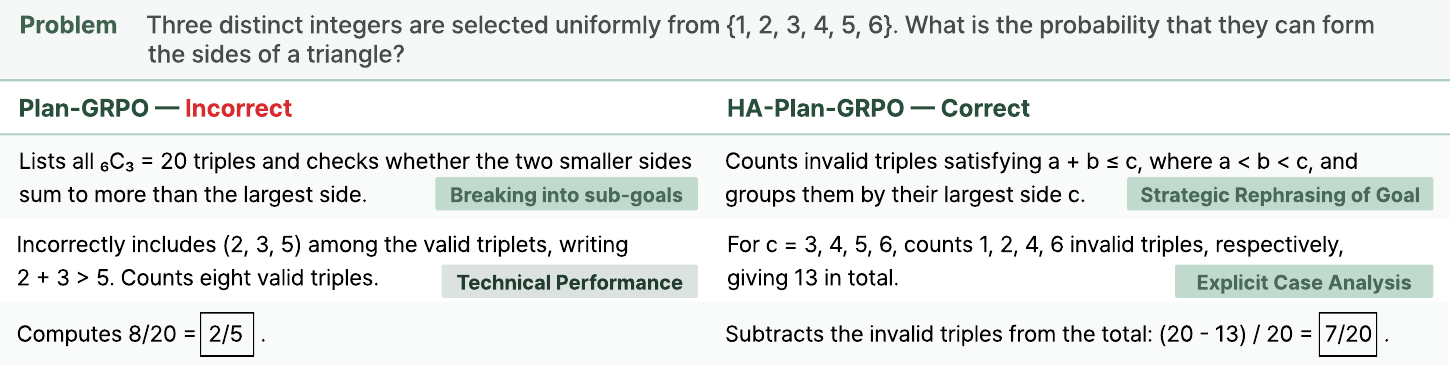}
\caption{
Paraphrased evaluation-time solutions from Qwen3.5-4B
checkpoints trained with Plan-GRPO and HA-Plan-GRPO.
Both models generate the response to the same zero-shot CoT prompt without
heuristic guidance at inference.
}
\label{fig:ha-grpo-triangle}
\end{figure*}

\subsection{Task Performance}
\label{sec:ha-grpo-performance}

\paragraph{Setup}
We compare two variants of planning-based RLVR using Group Relative Policy Optimization (GRPO)~\citep{shao2024deepseekmath}.
Plan-GRPO prompts the model to state its goal and subgoals before solving, following prior planning-based reasoning work~\citep{jiao2024learning}.
Heuristic-Augmented Plan-GRPO (HA-Plan-GRPO) additionally provides the eleven heuristic families in \texttt{SHAPE} and prompts the model to select a relevant heuristic and explain its applicability before each major reasoning step.
We train both variants on the MATH training split~\citep{hendrycks2021measuring} using three models: DeepSeek-Distill-Llama-8B, Qwen3-1.7B-Base, and Qwen3.5-4B-Base.
Within each model, the variants share the training data, reward, verifier, optimizer, and training budget; only the training rollout prompt differs.
All checkpoints are evaluated using the same zero-shot CoT prompt for each dataset.
We report Avg@16 and Pass@16 on the original, simple, and hard MATH-Perturb test sets~\citep{huang2025math}, MATH500, and Minerva~\citep{lewkowycz2022minerva}.
Results for trained variants are averaged across three training seeds.
Full training and evaluation details, including prompts, are provided in Appendix~\ref{app:experiment_details}.

\paragraph{Results}
Explicit mathematical heuristic guidance improves RLVR beyond planning alone.
HA-Plan-GRPO achieves higher mean Avg@16 and Pass@16 than Plan-GRPO across all three models (Table~\ref{tab:ha_grpo_mathp}).
Across the fifteen model--evaluation-set comparisons, it improves Avg@16 in fourteen and Pass@16 in fourteen.
Averaged equally over the five evaluation sets, the gains range from $1.1$ to $7.0$ percentage points in Avg@16 and from $0.4$ to $2.4$ percentage points in Pass@16.
The largest gains occur for DeepSeek-Distill-Llama-8B, where mean Avg@16 increases from $42.8\%$ to $49.8\%$ and mean Pass@16 from $69.7\%$ to $72.1\%$.
These gains are obtained by modifying the training rollout prompt while retaining the same training data, outcome-based reward, and optimization setup.
They persist under a shared zero-shot CoT evaluation prompt, demonstrating that the mathematical heuristic vocabulary of \texttt{SHAPE} provides practical guidance for improving post-training.

\subsection{Qualitative Analysis}
\label{sec:ha-grpo-qualitative}

Figure~\ref{fig:ha-grpo-triangle} contrasts how heuristics organize two responses to the same triangle-counting task. 
Both Qwen3.5-4B checkpoints receive the same zero-shot CoT prompt without planning or heuristic guidance. 
The Plan-GRPO response uses Breaking into Sub-goals to outline the counting procedure, then proceeds through Technical Performance by checking all twenty triples individually. 
Its sub-goals specify what to compute, but do not further organize the candidates into specific cases. 
During these individual checks, it incorrectly accepts $(2,3,5)$ despite $2+3=5$, inflating the count and producing the incorrect probability $2/5$.
The HA-Plan-GRPO, in contrast, combines Strategic Rephrasing of Goal with Explicit Case Analysis. 
It reformulates the task as counting invalid triples and splits them by their largest side, turning the count into a set of mutually exclusive subproblems. 
Within these cases, it correctly applies the failure condition $a+b\leq c$, including degenerate triples among the invalid cases, and obtains $7/20$. 
This example illustrates how heuristic-guided training can help models use mathematical heuristics such as case decomposition to organize counting, rather than relying on brute-force enumeration of all combinations.

\section{Conclusion}
\label{sec:conclusion}
We introduced \texttt{SHAPE}, a framework grounded in mathematics-education research that analyzes the mathematical actions expressed in CoT trajectories through a fixed, problem-independent heuristic taxonomy.
Analyses show that this shared vocabulary reveals differences in strategy use across superficially similar problems with different solution requirements.
We also find that successful RL solutions show substantial overlap with base-model profiles, while coverage varies across pairs.
Using the same taxonomy to guide planning during RLVR improves Avg@16 and Pass@16 over planning alone across three models under a shared evaluation prompt.
Our qualitative example illustrates how case decomposition and goal reformulation can organize the execution of a plan.
Together, these findings show how a common vocabulary for describing mathematical strategies can support both systematic analysis of LLM reasoning and improvements to post-training.

\newpage
\section*{Limitations}
\texttt{SHAPE} analyzes mathematical actions expressed in generated CoT traces and does not assess whether these traces faithfully reflect the internal reasoning process. 
Concepts from mathematics education provide a vocabulary for categorizing observable actions; their use does not imply that model-generated CoT is equivalent to human verbalized reasoning.
Also, the current framework operates on individual CoT trajectories and does not model reasoning distributed across multiple agents or developed through interactive dialogue. Applying \texttt{SHAPE} to these settings would require extending the annotation protocol to account for speaker roles, dialogue turns, and dependencies between contributions.

\section*{Ethical Considerations}
LLM-based annotation and RL post-training consume computational resources and energy, creating an environmental cost that may increase with scale. We do not quantify energy consumption or carbon emissions in this study. These costs should be considered when extending the framework to larger annotation or training workloads.

\newpage
\bibliography{custom}


\newpage
\appendix
\raggedbottom
\arrayrulecolor{black}
\captionsetup{position=bottom,hypcap=false}
\renewcommand{\dbltopfraction}{0.95}
\renewcommand{\dblfloatpagefraction}{0.45}
\makeatletter
\setlength{\@dblfptop}{0pt}
\makeatother

\newtcblisting{promptbox}[1]{
    enhanced,
    colback=blue!5,
    colframe=blue!50,
    title=\textbf{#1},
    fonttitle=\bfseries,
    boxrule=0.5pt,
    arc=2pt,
    width=\textwidth,
    listing only,
    listing options={
        basicstyle=\ttfamily\small,
        breaklines=true,
        columns=fullflexible,
        keepspaces=true,
        showstringspaces=false
    }
}

\section{Heuristic Taxonomy}
\label{app:taxonomy}

In this section, we provide the full label set used for \texttt{SHAPE} annotation. The taxonomy is designed as an operational coding scheme for LLM-generated mathematical CoT trajectories. Its purpose is to give the annotator model a stable vocabulary for identifying the local mathematical action expressed in each content unit.
The taxonomy is grounded in prior work on mathematical problem solving~\citep{polya1945solve,koichu2007heuristic,favier2022etude,posamentier2008problem}. We consolidate these sources into 11 top-level heuristic families. These families cover representation changes, reinterpretation, formalization, problem classification, simplification, case decomposition, contradiction, analogy, pattern exploration, backward reasoning, and verification. Some families are further divided into sublabels when a finer distinction is useful for annotation. For example, H3 separates the introduction of notation from structural augmentation, and H11 separates direct checking, alternative derivation, backtracking, sanity checking, generalization, and reflection on rigor.
Labels are assigned at the content-unit level. A unit receives a heuristic label when it performs a purposeful mathematical action that can shape the subsequent solution process. The labeling is multi-label: a single unit may both introduce a representation and check a constraint, or both explore a case and verify a claim. If a unit does not contribute a strategic mathematical action, it is assigned one of the non-heuristic labels N1--N4. This distinction is functional rather than lexical. A line containing equations may still be N2 if it only carries out routine computation after the strategy has already been fixed.
Table~\ref{tab:taxonomy} lists the heuristic labels and their source mappings. Table~\ref{tab:non_heuristics} lists the non-heuristic labels used to preserve non-strategic parts of the generated CoT trajectory.

\begin{table*}[p]
\centering
\small
\begin{tabular}
{@{}>{\raggedright\arraybackslash}p{5cm}>{\raggedright\arraybackslash}p{\dimexpr\textwidth-5cm-2\tabcolsep\relax}@{}}
\toprule
\textbf{Strategy Name} & \textbf{Description \& Sources} \\
\midrule
    H1.\ Changing the register of semiotic representation & This strategy involves translating the problem's representation from one semiotic register to another. It includes converting between natural language, algebraic, geometric, and visual representations to facilitate understanding or solving. \newline \textbf{Sources:} Creating a model (K); Change the semiotic representation register (Changer de registre de représentation sémiotique) (F) \\ [0.4em]
    
    H2.\ Cognitive Reinterpretation & This involves changing the way an object or property in the problem is interpreted. It redefines the identity or attributes of an element in a way different from the initial presentation, without necessarily changing the register. \newline \textbf{Sources:} Consider another way of interpreting the problem's objects (Envisager une autre façon d'interpréter les objets du problème) (F); Adopting a different point of view (P\&K) \\ [0.4em]
    
    H3.\ Introduce Symbolic Representation, Formalization, and Structural Augmentation, including... & \\ [0.4em]
    
    H3a.\ Introduce Symbolic Representation and Formalization & The act of introducing new variables, labeling unknowns, or performing substitutions to make ambiguous targets operationally manageable.  \newline \textbf{Sources:} Notation; Setting up equations (P); Creating a model (K); Introduce names or notations (Introduire des noms ou des notations) (F) \\ [0.4em]
    
    H3b.\ Structural Augmentation & Constructing auxiliary objects, lemmas, or entirely new mathematical frameworks that are not present in the original problem. This is a creative addition to the problem space, such as drawing auxiliary lines, defining new functions, or shifting the problem into a new structural representation \newline \textbf{Sources:} Auxiliary elements, Lemma (P); Introducing an auxiliary element (K); Introduce auxiliary elements (Introduire des éléments auxiliaires) (F) \\ [0.4em]
    
    H4.\ Problem Classification / Rephrase the Problem and Goal, including... & \\ [0.4em]
    
    H4a.\ Problem Categorization / Strategic Rephrasing of Goal / Breaking into Sub-goals & Explicitly stating the problem type, identifying applicable solution methods, or reformulating the main goal in clearer mathematical terms. \newline  \textbf{Sources:}What is the unknown? (P); Reformuler le problème (Reformulate the problem) (F) \\ [0.4em]
    
    H4b.\ Filtering Constraints & Strategically identifying the most essential constraints or conditions that guide the upcoming solution approach. \newline \textbf{Sources:}Separate the various parts of the condition (P); Exploring a particular datum (K) \\ [0.4em]
    
    H5.\ Wishful Thinking (Simplify / Reduce the Problem and Conditions) & Temporarily modifying the problem to a simpler version to gain insight, verify formulas, or explore solution strategies. \newline \textbf{Sources:} If you cannot solve the proposed problem (P); Reduce the problem to a simpler one (Réduire le problème à un problème plus simple) (F)  \\ [0.4em]
\bottomrule
\end{tabular}
\end{table*}

\begin{table*}[p]
\centering
\small
\begin{tabular}
{@{}>{\raggedright\arraybackslash}p{5cm}>{\raggedright\arraybackslash}p{\dimexpr\textwidth-5cm-2\tabcolsep\relax}@{}}
\toprule
\textbf{Strategy Name} & \textbf{Description \& Sources} \\
\midrule
    H6.\ Explicit Case Analysis, Decompose into Subproblems & Logically decomposing the problem into distinct cases, non-overlapping subsets, or sub-problems that, when combined, yield the full solution. The cases should ideally be exhaustive and mutually exclusive. \newline \textbf{Sources:} Decomposing and recombining (P); Décomposer le domaine du problème et travailler cas par cas (Decompose the problem domain and work case by case) (F) \\ [0.4em]
    
    H7.\ Arguing by contradiction & A proof strategy where the negation of the proposition is assumed to derive a contradiction, thereby proving the original statement.  \newline \textbf{Sources:} Reductio ad absurdum and indirect proof (P); Arguing by contradiction (K) \\ [0.4em]
    
    H8.\ Analogy and Presenting Related Theorems, including...  & \\ [0.4em]
    
    H8a.\ Analogy & Recalling previously solved problems, known methods, or applying a recently established logical procedure to a new target within the same problem. This involves recognizing structural similarities and transferring a strategy from one context (or one part of the equation) to another. \newline \textbf{Sources:} Have you seen it before?; Do you know a related problem? (P); Activating a previous experience (K); Make a connection with a previously encountered problem (Faire le lien avec un problème déjà rencontré) (F); Solving a simpler analogous problem (P\&K) \\ [0.4em]
    
    H8b.\ Presenting Related Theorems, Tools, or Properties & Introducing specific mathematical theorems, formulas, identities, or properties that are not provided in the problem statement but are necessary to advance the solution. \newline  \textbf{Sources:} Connect with a mathematical tool (theorem, property) (Faire le lien avec un outil mathématique (théorème, propriété)) (F) \\ [0.4em]
    
    H9.\ Experimental and Pattern Exploration, including... & \\ [0.4em]
    
    H9a.\ Exploring particular cases or numbers & Plugging in specific values, extreme/boundary values, or limits to discover patterns, build intuition, or verify feasibility.\newline \textbf{Sources:} Specialization (P); Partial Induction (K); Explore a specific piece of data (Explorer une donnée particulière) (F); Finding a Pattern (P\&K)\\ [0.4em]
    
    H9b.\ Exploration of symmetry & Identifying and exploiting mathematical or structural symmetry to reduce the solution space or simplify computation. \newline \textbf{Sources:} Symmetry (P); Exploration of Symmetry (K); Exploit symmetry properties (Exploiter les propriétés de symétrie) (F) \\ [0.4em]
    
    H10.\ Thinking from the end to the beginning (Working backward) & Starting from the desired conclusion (target goal) and working logical steps backward to reach the known premises or to determine what would be sufficient to prove. \newline \textbf{Sources:} Working backwards (P); Thinking Backward (K); Working backward (Travailler à reculons) (F); Working Backwards (P\&K)\\ [0.4em]
    
    H11.\ Verification and Looking Back, including... & \\ [0.4em]
    
    H11a.\ Re-solving \& Checking the Argument & Re-performing the same logical steps or calculations without a strategic change, or conducting a direct manual check of elements to verify a previous claim. \newline \textbf{Sources:} Can you check the result? (P); Local Self-evaluating (K) \\ [0.4em]
    
    H11b.\ Deriving the Result Differently & Solving the same problem or sub-goal using a structurally different mathematical method to provide independent confirmation. \newline \textbf{Sources:} Can you derive the result differently? (P); Local Self-evaluating (K) \\ [0.4em]
    
    H11c.\ Backtracking for self-verification & Realizing an error, finding a flaw in an assumption, or recognizing that the current approach is not working, and revising the direction. \newline \textbf{Sources:} Backtracking (Retour arrière) (F) \\ [0.4em]
    
    H11d.\ Checking the Result / Sanity Check / Progress Review & Broadly covers any reflection on whether the solution is on the right track or checking feasibility. \newline \textbf{Sources:} Can you check the result?; Test by dimension (P); Local Self-evaluating (K) \\ [0.4em]
    
    H11e.\ Generalization \& Corollary & Extending the result to broader cases. Identifying general principles from specific solutions. \newline \textbf{Sources:} Wisdom of proverbs (P); Generalization (K) \\ [0.4em]
    
    H11f.\ Reflect on Rigor \& Wisdom & Evaluating the efficiency of the solution strategy, questioning the rigor, or meta-reflecting on definitions/rules. \newline \textbf{Sources:} Why proofs? (P) \\
\bottomrule
\end{tabular}
\caption{Taxonomy of Problem-Solving Strategies. This table summarizes the code, strategy name, detailed description, and theoretical sources associated with each heuristic. Here, P = \citet{polya1945solve}, K = \citet{koichu2007heuristic}, F = \citet{favier2022etude}, and P\&K = \citet{posamentier2008problem}.}\label{tab:taxonomy}
\end{table*}

\begin{table}[!t]
\centering
\small
\begin{tabularx}{\linewidth}{@{}>{\raggedright\arraybackslash}p{2.05cm}>{\raggedright\arraybackslash}X@{}}
\toprule
Name & Description \\
\midrule
N1.\ Literal Repetition & Merely repeating the problem statement or giving a generic opening/closing without mathematical substance or added insight. \\
N2.\ Technical Performance & Routine calculations or simple algebraic manipulations that do not involve strategic planning or new insight; ``doing the math'' after the plan is set. \\
N3.\ Alien Statements & Unrelated thoughts or off-topic remarks. \\
N4.\ Answer & Stating the final answer. \\
\bottomrule
\end{tabularx}
\caption{Non-heuristic categories capturing reasoning steps that do not contribute strategic insight, including repetition, routine computation, irrelevant statements, and final answer reporting.}
\label{tab:non_heuristics}
\end{table}

\section{Gold Heuristic Set Constructions}
\label{app:construction}
\label{app:gold_protocol}

The gold set is intended as an adjudicated reference for validating the \texttt{SHAPE} annotation pipeline, not as an assumption that heuristic interpretation is observer-independent. Following prior work in mathematical problem-solving analysis, we treat annotation as an interpretive coding task whose reliability comes from explicit coding rules, examples, and consensus adjudication.

\paragraph{Trajectory selection.}
To broaden the range of reasoning traces represented in the gold annotation set, we deliberately varied three factors: seed problems, perturbation conditions, and model configurations. 
We selected six seed problems from MATH-Perturb, with problem IDs 12, 968, 1694, 1958, 2423, and 3156. 
For each seed problem, we included both the original problem and its hard perturbation. 
We collected CoT trajectories for each version from four models spanning different reasoning modes and model sizes: Qwen3-30B-A3B-Instruct, Qwen3-30B-A3B-Thinking, Qwen3-8B, and Nemotron-Cascade-8B. Including every problem--condition--model combination yielded $6 \times 2 \times 4 = 48$ CoT trajectories. 
This crossed design was intended to broaden coverage of heuristic usage across problem instances, original and structurally perturbed settings, and model configurations.

\paragraph{Content-unit segmentation.}
We segmented each CoT trajectory into content units following the protocol-coding approach of Koichu et al.~\citep{koichu2007heuristic}. A content unit is the largest unbroken span of the trajectory that supports a particular heuristic interpretation. In practice, a content unit may consist of a few words, one sentence, or several sentences, depending on whether the span expresses a single problem-solving move. We introduced a new unit boundary when the model changed the local mathematical goal, shifted to a different action, moved from planning to execution, moved from computation to checking, or began a new attempt at representing or transforming the problem.

\paragraph{Consensus coding.}
Four authors participated in the annotation process, including a graduate researcher in mathematics education. The trajectories were coded at the content-unit level. Disagreements were resolved through discussion, following the interpretive tradition in mathematical problem-solving research, where coding decisions are stabilized through comparison and adjudication rather than treated as purely mechanical labels. During discussion, annotators considered the problem statement, the local content unit, and the surrounding trajectory context. The final annotation was retained only after the annotators reached a shared interpretation of the mathematical role of the unit.
Our procedure follows Koichu et al.~\citep{koichu2007heuristic}, who analyze thinking-aloud transcripts by segmenting them into content units and coding each unit according to its heuristic interpretation. In their protocol, content units are defined as the largest unbroken parts of the transcript that bear a particular heuristic interpretation, and disagreements are resolved by comparing coders' interpretations and adjudicating ambiguous cases. We adapt this procedure from human think-aloud data to LLM-generated CoT trajectories.

\paragraph{Borderline annotation example.}
One representative disagreement occurred in a trajectory generated by Qwen3-30B-A3B-Thinking on the hard perturbation of MATH-Perturb problem 1694. In this trajectory, the model tested its counting method on smaller ranges:
\begin{quote}
``But just to make 100\% sure, let's compute the total numbers with $\gcd(n,28)=2$ in a small range and see if our method works. Take $n>2$, $n<10$: numbers $3$--$9$. ... So only $n=6$ has $\gcd=2$. Using our method ... Count=1. ... Another test: $n>6$, $n<20$.''
\end{quote}
Annotators initially differed on whether this unit should be treated as H9a, H11b, or H5. The H9a interpretation comes from the direct exploration of particular numerical cases. The H11b interpretation comes from the unit's role in the trajectory: the model has already derived a general counting method and uses these examples to check whether that method works. The H5 interpretation comes from reducing the original range to smaller test ranges. In adjudication, we retained all three labels because the unit performs all three roles: it verifies the solution by testing particular cases in a simplified version of the original problem.

\section{Annotator Model Details}
\label{app:annotator_prompts}

\label{app:annotator-model-prompts}

This appendix reports the prompt skeletons used by the automated \texttt{SHAPE}
annotation pipeline. To avoid duplicating long coding manuals in the paper,
we omit the full guidebook text from the printed prompt listings. In the actual
annotation runs, guidebook placeholders were expanded verbatim at runtime.

Specifically, \texttt{\{HEURISTICS\_\allowbreak{}GUIDEBOOK\}} was replaced by the full
heuristic annotation guidebook. The guidebook, together with the prompt-building
code and exact runtime templates, is included in the supplementary code release.
Thus, each annotator call is defined by the prompt
skeleton shown below plus the corresponding guidebook file in the supplementary
materials.

For readability, the prompt listings below retain the guidebook placeholders
rather than expanding them in full.

\subsection{Content-Unit Segmentation Prompt}

\begin{figure*}[!t]
\centering
\begin{promptbox}{Content-Unit Segmentation Prompt}
# Role
You are an expert researcher in mathematics education and cognitive psychology.
Your task is to segment a student's or AI model's Chain-of-Thought (CoT) transcript
into Content Units using the provided heuristic guidebook.

# Task Definition
Your primary goal is to segment the Chain-of-Thought transcript into Content Units
based on shifts in the problem-solving strategy.

A Content Unit is a single Plan-Execution Cycle:
A cohesive block of thought where the solver formulates a specific strategic
intent and immediately carries it out.

Segmentation Philosophy:
- Do not segment every sentence.
- Do not segment based only on pauses, filler words, or discourse markers.
- Segment only when the active strategy or underlying mathematical intent changes.
  For example, split when the solver switches from analyzing the problem to
  experimenting with numerical cases.

Crucial Segmentation Rules:
1. Coverage is mandatory: every sentence in the target range must be assigned to a unit.
2. Merge a plan and its immediate execution into one unit.
3. Use heuristic codes as justification for the segmentation boundary.
4. Start a new segment only when the active strategy changes.
5. Group consecutive sentences with the same strategic role into a single unit.
6. Merge weak or monitoring sentences such as "Okay", "Wait", or "Let me check"
   with neighboring substantive units.
7. Do not output N/A, empty codes, or non-heuristic labels.
8. Do not create single-sentence chunks for weak monitoring or transition sentences,
   especially when they are shorter than about 15 words.

# Reference: Heuristic Strategies
Use the heuristic taxonomy from the guidebook. The valid top-level heuristic families are:

\end{promptbox}
\end{figure*}

\begin{figure*}[!t]
\centering
\begin{promptbox}{Content-Unit Segmentation Prompt (continued)}
H1:  Changing the register of semiotic representation
H2:  Cognitive reinterpretation
H3:  Introduce symbolic representation, formalization, and structural augmentation
H4:  Problem classification / rephrasing the problem and goal
H5:  Wishful thinking: simplifying or reducing the problem and conditions
H6:  Explicit case analysis / decomposing into subproblems
H7:  Arguing by contradiction
H8:  Analogy and presenting related theorems, tools, or properties
H9:  Experimental and pattern exploration
H10: Thinking from the end to the beginning / working backward
H11: Verification and looking back

# Critical Valid-Code Rule
The only valid codes in this segmentation stage are H1-H11.
Never use N1-N4, "Non-Heuristic", "Technical Performance", or any other non-H label.

# In-Context Example
Problem:
Find the least positive four-digit solution r of the congruence
r^2 + 4r + 4 = r^2 + 2r + 1 (mod 55).

Input CoT:
{SEGMENTATION_IN_CONTEXT_EXAMPLE_COT}

Output Segmentation:
{
  "chunks": [
    {"start_index": 1,  "end_index": 9,  "codes": ["H4"],        "reasoning": "..."},
    {"start_index": 10, "end_index": 15, "codes": ["H11","H8"],  "reasoning": "..."},
    {"start_index": 16, "end_index": 27, "codes": ["H8"],        "reasoning": "..."},
    {"start_index": 28, "end_index": 31, "codes": ["H11","H8"],  "reasoning": "..."},
    {"start_index": 32, "end_index": 46, "codes": ["H11"],       "reasoning": "..."},
    {"start_index": 47, "end_index": 57, "codes": ["H4"],        "reasoning": "..."}
  ]
}

# Instruction
Segment the following User Input, a Chain-of-Thought trace, into Content Units
based on shifts in the problem-solving strategy.

Important: Data minimization
- Output start_index and end_index only.
- Do not output the text content of the chunks.

\end{promptbox}
\end{figure*}

\begin{figure*}[!t]
\centering
\begin{promptbox}{Content-Unit Segmentation Prompt (continued)}
# User Input
{INDEXED_SENTENCE_WINDOW}

Important:
1. You are provided with a window of sentences, each indexed as [1], [2], ...
2. Segment from index {TARGET_START_INDEX} up to roughly index {TARGET_END_INDEX}.
3. Start grouping from index {TARGET_START_INDEX}. Do not include earlier indices.
4. If a logical unit extends past {TARGET_END_INDEX}, stop before it and leave it
   for the next batch.
5. Group consecutive sentences with the same strategic role into a single chunk.
6. Never output a chunk containing only a weak monitoring sentence.
7. Every chunk must have at least one valid H1-H11 code.
8. Do not create single-sentence chunks for weak monitoring, confirmation, or transition.
9. Valid codes are H1-H11 only. Never use N1-N4 or any non-H prefix.
10. Output strictly as JSON with a "chunks" key:
    {
      "chunks": [
        {
          "start_index": 1,
          "end_index": 5,
          "codes": ["H4"],
          "reasoning": "Brief explanation."
        }
      ]
    }
11. Each item must include start_index, end_index, codes, and reasoning.
12. Keep reasoning concise: 1-2 sentences.
13. Do not quote long blocks of the input text.
14. Use single quotation marks inside reasoning if quotation is needed.
\end{promptbox}
\end{figure*}

\paragraph{Runtime placeholders.}
\texttt{\{INDEXED\_\allowbreak{}SENTENCE\_\allowbreak{}WINDOW\}} is the sliding window of indexed sentences.
\texttt{\{TARGET\_\allowbreak{}START\_\allowbreak{}INDEX\}} is the first sentence index to segment in the current batch.
\texttt{\{TARGET\_\allowbreak{}END\_\allowbreak{}INDEX\}} is the approximate end index for the current batch.
\texttt{\{SEGMENTATION\_\allowbreak{}IN\_\allowbreak{}CONTEXT\_\allowbreak{}EXAMPLE\_\allowbreak{}COT\}} is the fixed in-context example used for calibration.

\subsection{Heuristic Tagging Prompt}

\begin{figure*}[!t]
\centering
\begin{promptbox}{Heuristic Tagging Prompt}
In this project, we analyze the reasoning process of Large Reasoning Models (LRMs)
by identifying heuristic strategies used during mathematical problem solving.
Given one content unit from the model response, annotate it with heuristic or
non-heuristic codes from the guidebook.

The code set includes:
- H1-H11: mathematical heuristic strategies.
- N1-N4: non-heuristic categories, used only when no heuristic strategy is present.

Important Annotation Rules:
1. Use sub-codes when available. For categories with sub-codes, such as H4a/H4b,
   H9a/H9b, or H11a-H11f, use the specific sub-code if it applies. Use the parent
   code only as a fallback when no sub-code fits.
2. Multi-tagging is allowed. A content unit may receive multiple codes if multiple
   strategies are present.
3. Use non-heuristic codes only when no heuristic is present. If the unit contains
   even one H1-H11 action, output H codes only.
4. Cite evidence. For each code, quote the specific phrase or sentence from the
   current chunk that supports the annotation.

The [Guidebook] section provides the full definitions of the codes.
The [Math Problem] section provides the original math problem.
The [Previous Context] section provides recent chunks without previous tags.
The [Current Chunk] section provides the chunk to annotate.
The [Format] section specifies the required output format.

[Guidebook]
{HEURISTICS_GUIDEBOOK}
[End of Guidebook]

[Math Problem]
{PROBLEM_TEXT}
[End of Math Problem]

[Previous Context]
Recent chunks, text only:
Chunk {i-3}: {prev_chunk_text_i_minus_3_truncated_200chars}
Chunk {i-2}: {prev_chunk_text_i_minus_2_truncated_200chars}
Chunk {i-1}: {prev_chunk_text_i_minus_1_truncated_200chars}
Chunk {i}:   {prev_chunk_text_i_truncated_200chars}
[End of Previous Context]

\end{promptbox}
\end{figure*}

\begin{figure*}[!t]
\centering
\begin{promptbox}{Heuristic Tagging Prompt (continued)}
[Current Chunk]
Chunk to annotate:
{CURRENT_CHUNK_TEXT}
[End of Current Chunk]

[Format]
Output JSON exactly in the following structure:
{
  "annotations": [
    {
      "code": "H4",
      "evidence": "Direct quote from the chunk serving as evidence.",
      "reasoning": "Brief explanation of why this code applies."
    }
  ]
}

The "annotations" field must be a list.
Each object must contain exactly one code.

# Heuristic-First Decision Rule
Before annotating, follow this decision tree:
1. First, scan carefully for any heuristic activity, H1-H11.
2. If you find even one heuristic, tag with H codes only.
3. Only if there is no heuristic activity, use N codes.

Common mistakes to avoid:
- Do not tag modular arithmetic operations as N2 when they introduce or use a
  mathematical representation or tool. Use H1 or H8 when appropriate.
- Do not tag finding an inverse or invoking a theorem as N2. Use H8 when it
  functions as a mathematical tool.
- Do not tag strategic substitutions, such as "Let u = x^2", as N2. Use H3.
- Do not tag problem-to-equation conversion as N2. Use H1 or H3.

N2 is only for routine arithmetic or algebra after the strategy has already been set,
such as "2+3=5" or "2x+3x=5x".

Now annotate the current chunk using the heuristic-first approach.
\end{promptbox}
\end{figure*}

\paragraph{Runtime placeholders.}
\texttt{\{HEURISTICS\_\allowbreak{}GUIDEBOOK\}} is the full heuristic guidebook.
\texttt{\{PROBLEM\_\allowbreak{}TEXT\}} is the original math problem.
\texttt{\{prev\_\allowbreak{}chunk\_\allowbreak{}text\_\allowbreak{}*\}} are up to four recent prior chunks, truncated to 200 characters.
If no previous chunks exist, this field is replaced with ``There are no previous chunks.''
\texttt{\{CURRENT\_\allowbreak{}CHUNK\_\allowbreak{}TEXT\}} is the content unit being annotated.

\section{Sensitivity to Annotator Choice}
\label{app:annotator_sensitivity}

\paragraph{Matched trajectories.}
We compare Qwen3.5-27B and Grok-4.1-Fast on the same CoT trajectories to assess whether the full-trajectory perturbation contrast in \S\ref{sec:mode-seeking-hard} depends on the annotator.
The comparison covers four solvers: Nemotron-Cascade-8B, Olmo-3-7B-Think-RLVR, Qwen3-8B, and Qwen3-32B, each evaluated on 115 original, simple, and hard problems (1,380 trajectories).
We retain 1,370 trajectories with complete annotations from both sources and identical concatenated CoT text after removing whitespace; two missing Grok annotations and eight text mismatches are excluded.
Of the retained pairs, 1,033 have identical content-unit boundaries and texts, while the remaining 337 Qwen3-8B pairs use different segmentations of the same text.
For both sources, we construct $u$ using the eleven heuristic families in Eq.~\ref{eq:heuristic-frequency}, collapsing sublabels to their parent family, counting each family once per unit, and excluding non-heuristic labels.

\paragraph{Trajectory-level consistency.}
Across the 1,370 matched trajectories, the cosine similarity between Qwen- and Grok-derived heuristic-frequency distributions is $0.956 \pm 0.034$ (mean $\pm$ standard deviation).
This measures agreement between the two automated annotations of the same text; the larger comparison set has no additional human reference labels.

\paragraph{Stability of the perturbation contrast.}
To test the downstream comparison directly, for annotator $a$ and seed problem $i$ we compute
\begin{equation}
\begin{aligned}
C_i^{(a)} &= D_{\mathrm{JS}}(u_{i,O}^{(a)},u_{i,H}^{(a)})\\
&\quad - D_{\mathrm{JS}}(u_{i,O}^{(a)},u_{i,S}^{(a)}),
\end{aligned}
\label{eq:annotator-perturbation-contrast}
\end{equation}
where O, S, and H denote original, simple, and hard conditions, respectively, and $D_{\mathrm{JS}}$ is the base-2 Jensen--Shannon divergence computed from full trajectories.
We use the same valid O/S/H triples for both annotators and estimate 95\% percentile confidence intervals from 10,000 bootstrap resamples of seed problems, retaining both annotation sources together in each resample.
Table~\ref{tab:annotator_sensitivity_contrast} reports the resulting mean contrasts.
For every tested solver, the mean contrast is positive under both annotators, with all eight confidence intervals above zero.
Changing the annotator changes the mean contrast by at most $0.0041$.
Thus, the direction of the full-trajectory hard-versus-simple difference is preserved across the two annotators for these four solvers.

\begin{table}[t]
\centering
\small
\setlength{\tabcolsep}{2.5pt}
\begin{tabular}{@{}lrrr@{}}
\toprule
Solver & $n$ & Qwen & Grok \\
\midrule
\shortstack[l]{Nemotron-Cascade\\8B} & 115 &
\shortstack{$0.019$\\$[0.008,0.029]$} &
\shortstack{$0.023$\\$[0.014,0.032]$} \\
\addlinespace[3pt]
\shortstack[l]{Olmo-3-7B\\Think-RLVR} & 115 &
\shortstack{$0.022$\\$[0.012,0.033]$} &
\shortstack{$0.023$\\$[0.015,0.032]$} \\
\addlinespace[3pt]
Qwen3-8B & 107 &
\shortstack{$0.024$\\$[0.014,0.035]$} &
\shortstack{$0.022$\\$[0.014,0.032]$} \\
\addlinespace[3pt]
Qwen3-32B & 113 &
\shortstack{$0.021$\\$[0.009,0.032]$} &
\shortstack{$0.022$\\$[0.014,0.031]$} \\
\bottomrule
\end{tabular}
\caption{Mean full-trajectory hard-minus-simple JSD contrasts under Qwen3.5-27B and Grok-4.1-Fast. Brackets contain problem-bootstrap 95\% confidence intervals; $n$ is the number of seed problems with complete matched O/S/H triples.}
\label{tab:annotator_sensitivity_contrast}
\end{table}

These results establish consistency of this aggregate contrast, rather than equality of individual labels or absolute divergence estimates.
They do not establish human-reference accuracy on the larger set or annotator robustness of the Density/Coverage and training analyses.

\section{Robustness of Density and Coverage}
\label{app:dc_robustness}

\paragraph{Representation and scope.}
All comparisons use the same representation, sample-selection rule, and Density/Coverage implementation as Table~\ref{tab:dc}.
We use cosine distance in the original annotation-frequency space.
For each model pair, we retain problem--condition variants with at least one successful annotation from each model and pool all available successful trajectories, without downsampling.
Reference neighborhoods are closed: a target is included when its distance is at most the distance to the $k$-th nearest reference trajectory, excluding the reference itself.

\paragraph{Neighborhood size.}
Table~\ref{tab:dc_k_sensitivity} varies $k\in\{3,5,10\}$ while holding each pair's trajectories fixed.
It includes all six comparisons in Table~\ref{tab:dc} and an additional Olmo-3-7B--Think-RLVR comparison using the same selection rule.
The $k=3$ entries for the six main-table pairs are identical to Table~\ref{tab:dc}.

\begin{table*}[t]
\centering
\small
\setlength{\tabcolsep}{4pt}
\resizebox{\linewidth}{!}{%
\begin{tabular}{llrrrrrrrr}
\toprule
Reference & Target & $N_{\text{ref}}$ & $N_{\text{target}}$
& \multicolumn{2}{c}{$k=3$}
& \multicolumn{2}{c}{$k=5$}
& \multicolumn{2}{c}{$k=10$} \\
\cmidrule(lr){5-6}
\cmidrule(lr){7-8}
\cmidrule(lr){9-10}
& & & & Density & Coverage
& Density & Coverage
& Density & Coverage \\
\midrule
\multicolumn{10}{l}{\textit{Post-trained}} \\
Qwen3-1.7B-Base & Qwen3-1.7B-GRPO
& 834 & 886
& 1.219 & 0.871
& 1.175 & 0.952
& 1.157 & 0.999 \\
Olmo-3-7B & Olmo-3-7B-Think-RL-Zero
& 1229 & 1307
& 1.250 & 0.707
& 1.272 & 0.806
& 1.272 & 0.897 \\
Qwen2.5-7B & Qwen2.5-7B-SimpleRL-Zoo
& 1280 & 1294
& 1.511 & 0.888
& 1.339 & 0.964
& 1.215 & 0.989 \\
DeepSeek-Math-7B-Base & DeepSeek-Math-7B-SimpleRL-Zoo
& 463 & 500
& 1.525 & 0.950
& 1.338 & 0.989
& 1.248 & 1.000 \\
\midrule
\multicolumn{10}{l}{\textit{Cross-model baselines}} \\
Qwen2.5-7B & DeepSeek-Math-7B-Base
& 796 & 493
& 0.594 & 0.348
& 0.557 & 0.438
& 0.488 & 0.568 \\
Olmo-3-7B & Qwen3-1.7B-Base
& 71 & 66
& 0.525 & 0.437
& 0.527 & 0.479
& 0.588 & 0.690 \\
\midrule
\multicolumn{10}{l}{\textit{Additional post-trained comparison}} \\
Olmo-3-7B & Olmo-3-7B-Think-RLVR
& 1507 & 1600
& 1.032 & 0.531
& 1.024 & 0.619
& 1.014 & 0.721 \\
\bottomrule
\end{tabular}%
}
\caption{Neighborhood-size sensitivity using the same successful trajectory pools and closed cosine neighborhoods as Table~\ref{tab:dc}. $N_{\text{ref}}$ and $N_{\text{target}}$ are fixed across $k$; Density and Coverage are reported separately. The final row adds the Olmo Think-RLVR checkpoint.}
\label{tab:dc_k_sensitivity}
\end{table*}
\section{Experiment Details}
\label{app:experiment_details}

This section provides experimental details for the experiments described in \S\ref{sec:mode-seeking}, and \S\ref{sec:ha-grpo-training}. We use 2$\times$NVIDIA B200 GPUs for GRPO training.

\subsection{Prompts}
\label{sec:prompts}

We use the following planning prompt templates for all GRPO training. The placeholder \texttt{\{problem\}} is replaced with problems in the training dataset.

\begin{figure*}[!t]
\centering
\begin{tcolorbox}[
    enhanced,
    width=\textwidth,
    colback=blue!5,
    colframe=blue!50,
    title=\textbf{Planning Prompt Template},
    fonttitle=\bfseries,
    boxrule=0.5pt,
    arc=2pt,
]
\small

\texttt{\{problem\}}
\\[1em]
Solve the problem step by step using explicit planning.
\\[0.5em]
Before solving, write a short plan that states the goal and the sub-goals or intermediate results you expect to need. Then carry out the plan one step at a time. The plan should guide the next action, not merely describe it afterward.
\\[0.5em]
Format:
\\[0.5em]
[Plan]\\
Goal:\\
Sub-goals:
\\[0.5em]
[Solution]
\\[0.5em]
[Step 1]\\
Reasoning:
\\[0.5em]
[Step 2]\\
Reasoning:
\\[0.5em]
...
\\[0.5em]
[Final Answer]\\
Please reason step by step, and put your final answer within \textbackslash boxed\{\}

\vspace{0.5em}

\end{tcolorbox}
\end{figure*}

\begin{figure*}[!t]
\centering
\begin{tcolorbox}[
    enhanced,
    width=\textwidth,
    colback=blue!5,
    colframe=blue!50,
    title=\textbf{Heuristic-Augmented Planning Prompt Template},
    fonttitle=\bfseries,
    boxrule=0.5pt,
    arc=2pt,
]
\small
\texttt{\{problem\}}
\\[1em]
Solve the problem step by step using heuristic-guided planning.
\\[0.5em]
At each major reasoning step, first choose one mathematical heuristic, briefly explain why it is useful, and then perform the corresponding reasoning move. The heuristic should guide the next action, not merely describe it afterward.
\\[0.5em]
Use only heuristics that genuinely help solve the problem. The list is not a checklist; do not force irrelevant heuristics into the solution.
\\[0.5em]
Available heuristics:\\
1. Change the Representation\\
2. Reinterpret the Object\\
3. Introduce New Symbols or Structures\\
4. Restructure the Goal\\
5. Wishful Thinking: Simplify Temporarily\\
6. Divide into Cases\\
7. Argue by Contradiction\\
8. Draw on Mathematical Knowledge\\
9. Explore with Examples and Symmetry\\
10. Work Backward from the Goal\\
11. Monitor, Verify, and Look Back\\
\\[0.5em]
Format:
\\[0.5em]
[Plan]\\
Goal:\\
Potentially useful heuristics:
\\[0.5em]
[Solution]
\\[0.5em]
[Step 1]\\
Heuristic:\\
Why this heuristic applies:\\
Reasoning:
\\[0.5em]
[Step 2]\\
Heuristic:\\
Why this heuristic applies:\\
Reasoning:
\\[0.5em]
...
\\[0.5em]

[Final Answer]\\
Please reason step by step, and put your final answer within \textbackslash boxed\{\}

\vspace{0.5em}

\end{tcolorbox}
\end{figure*}

\subsection{Training Details}
\label{app:training_details}
Table~\ref{tab:training-hyperparameters} summarizes the training hyperparameters. We train for 200 steps on the MATH training splits for approximately 3--3.5 hours. We set the maximum generation length to 2048 for training efficiency. GRPO algorithm is implemented in \texttt{verl}~\citep{sheng2025hybridflow}

\begin{table}[!t]

\centering
\small
\setlength{\tabcolsep}{3pt}
\begin{tabularx}{\linewidth}{@{}>{\raggedright\arraybackslash}X>{\raggedright\arraybackslash}X@{}}
\toprule
\textbf{Parameter} & \textbf{Value} \\
\midrule
Training Batch Size & 32 \\
Rollouts per Prompt & 4 \\
Number of GPUs & 2$\times$B200\\
Optimizer & AdamW \\
Temperature & 1.0 \\
Top-$p$ & 1.0 \\
Top-$k$ & -1 \\
Max Response Length & 2048 \\
Learning Rate & $1\times 10^{-6}$ \\
Warmup Steps & 10 \\
Training Steps & 200 \\
\bottomrule
\end{tabularx}
\caption{Hyperparameters used for GRPO training.}
\label{tab:training-hyperparameters}
\end{table}
\subsection{Licenses and Terms of Use}
\label{app:licenses}

We use publicly released datasets, model checkpoints, APIs, and software libraries in accordance with their respective licenses and terms of use. All assets are used only for research purposes. The main existing assets used in this work are:

\begin{itemize}
    \item \textbf{MATH-Perturb}: used for perturbation-based reasoning analysis; we follow the dataset license and terms released by the authors.
    \item \textbf{Qwen models}: used for generation, annotation, and training experiments; we follow the Qwen model license and HuggingFace model terms.
    \item \textbf{OLMo models}: used for base and post-trained model comparisons; we follow the Olmo model license and release terms.
    \item \textbf{Nemotron models}: used for model analysis and annotation-data construction; we follow the NVIDIA model license and HuggingFace model terms.
    \item \textbf{Closed-source API models}: Grok, GPT, and Gemini models are used through their respective API access conditions and provider terms.
    \item \textbf{\texttt{verl}}: used as the RLVR implementation framework; we follow its open-source software license.
\end{itemize}

\end{document}